\documentclass[10pt,twocolumn,letterpaper]{article}

\usepackage{iccv}
\usepackage{times}
\usepackage{epsfig}
\usepackage{graphicx}
\usepackage{amsmath}
\usepackage{amssymb}
\usepackage{subcaption}
\usepackage{caption}
\usepackage[binary-units=true]{siunitx}
\usepackage{booktabs}
\usepackage{gensymb}
\usepackage{commath}
\usepackage{cite}  
\usepackage{array}
\usepackage{fancyhdr}
\usepackage{setspace} 
\newcolumntype{M}[1]{>{\centering\arraybackslash}m{#1}}

\usepackage[pagebackref=true,breaklinks=true,letterpaper=true,colorlinks,bookmarks=false]{hyperref}

\iccvfinalcopy 

\def\iccvPaperID{229} 

\ificcvfinal\fi

\newcommand{\PAR}[1]{\vskip4pt \noindent{\bf #1~}}

\newcommand{\highlightcolor}{black}

\begin{document}
\title{Image-based localization using LSTMs for structured feature correlation}

\author{
F. Walch$^{1,3}$ \quad
C. Hazirbas$^1$ \quad
L. Leal-Taix{\'e}$^1$ \quad
T. Sattler$^2$ \quad
S. Hilsenbeck$^3$ \quad
D. Cremers$^1$  \\
{$^1$Technical University of Munich \quad$^2$Department of Computer Science, ETH Z\"urich \quad $^3$NavVis} }

\maketitle
\thispagestyle{fancy}

\begin{abstract}
   In this work we propose a new CNN+LSTM architecture for camera pose regression for indoor and outdoor scenes. 
  CNNs allow us to learn suitable feature representations for localization that are robust against motion blur and illumination changes. We make use of LSTM units on the CNN output, which play the role of a structured dimensionality reduction on the feature vector, leading to drastic improvements in localization performance.
   We provide extensive quantitative comparison of CNN-based
   \textcolor{\highlightcolor}{and} SIFT-based localization methods, showing the weaknesses and strengths of each. 
   Furthermore, we present a new large-scale indoor \textcolor{\highlightcolor}{dataset} with accurate ground truth from a laser scanner. 
   Experimental results on both indoor and outdoor public datasets show our method outperforms existing deep architectures, and can localize images in hard conditions, \eg, in the presence of mostly textureless surfaces, where classic SIFT-based methods fail.
\end{abstract}

\vspace{-0.3cm}


\section{Introduction}
Being able to localize a vehicle or device by estimating a camera pose from an image is a fundamental requirement for many computer vision applications such as navigating autonomous vehicles \cite{Lim12CVPR}, mobile robotics and Augmented Reality \cite{Lynen2015RSS}, and Structure-from-Motion (SfM) \cite{schoenberger2016sfm}. 

Most state-of-the-art approaches \cite{Li12ECCV,Svarm2016PAMI,Zeisl2015ICCV,Sattler2016PAMI} rely on local features such as SIFT \cite{Lowe04IJCV} to solve the problem of image-based localization. 
Given a SfM model of a scene, where each 3D point is associated with the image features from which it was triangulated, one proceeds in two stages \cite{Li10ECCV,Sattler11ICCV}: 
(i) establishing 2D-3D matches between features extracted from the query image and 3D points in the SfM model via descriptor matching; 
(ii) using these correspondences to determine the camera pose, usually by employing a $n$-point solver \cite{Kukelova13ICCV} 
inside a RANSAC loop \cite{Fischler81CACM}. 
%
\textcolor{\highlightcolor}{P}ose estimation can only succeed if enough correct matches have been found in the first stage. 
Consequently, limitations of both the feature detector, \eg, motion blur or strong illumination changes, or the descriptor, \eg, due to strong viewpoint changes, will cause localization approaches to fail.

\textcolor{\highlightcolor}{Recently}, two approaches have tackled the problem of localization with end-to-end learning. PlaNet \cite{weyandeccv2016} formulates localization as a classification problem, where the current position is matched to the best position in the training set. While this approach is suitable for localization in extremely large environments, it only allows to recover position but not orientation and its accuracy is bounded by the spatial extent of the training samples.
More similar in spirit to our approach, PoseNet \cite{kendall2015posenet, kendall2016bayesianpose} formulates 6DoF pose estimation as a regression problem. \textcolor{\highlightcolor}{In this paper, we show that PoseNet is significantly less accurate than} state-of-the-art SIFT 
methods \cite{Li12ECCV,Svarm2016PAMI,Zeisl2015ICCV,Sattler2016PAMI} \textcolor{\highlightcolor}{and propose a novel network architecture that significantly outperforms PoseNet}.

\begin{figure}[t!]
    \centering
	\begin{subfigure}{0.5\linewidth}
	\centering
	\includegraphics[width=0.95\linewidth]{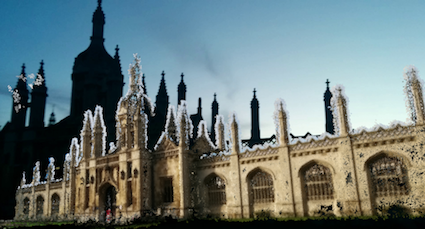}
	\caption{PoseNet result~\cite{kendall2015posenet}}
	\end{subfigure}\hfill
	\begin{subfigure}{0.5\linewidth}
	\centering
	\includegraphics[width=0.95\linewidth]{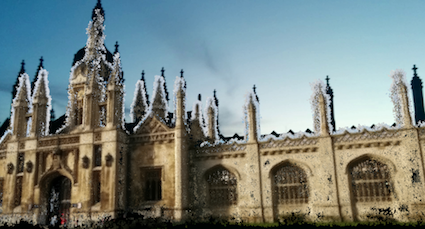}
	\caption{Our result}
	\end{subfigure}%
	\vspace{-6pt}%
    \caption{\textcolor{\highlightcolor}{Our approach achieves a}ccurate outdoor image-based localization 
    even in challenging lighting conditions where other deep architectures fail.}%
    \vspace{-6pt}%
\end{figure}

\subsection{Contributions}

In this paper, we propose to directly regress the camera pose from an input image. 
To do so, we leverage, on the one hand, Convolutional Neural Networks (CNNs) which allow us to learn 
suitable feature representations for localization that are more robust against motion blur and illumination changes.
As we can see from the PoseNet \cite{kendall2015posenet} results, regressing the pose after the high dimensional output of a FC layer is not optimal. Our intuition is that the high dimensionality of the FC output makes the network prone to overfitting to training data. PoseNet deals with this problem with careful dropout strategies.
We propose to make use of Long-Short Term Memory (LSTM) units \cite{lstm} on the FC output, which performs structured dimensionality reduction \textcolor{\highlightcolor}{and chooses} the most useful feature correlations for the task of pose estimation.
Overall, we improve localization accuracy by 32-37\% wrt. previous deep learning architectures \cite{kendall2015posenet,kendall2016bayesianpose}. 
Furthermore, 
we are the first to provide an extensive comparison with a state-of-the-art SIFT-based method \cite{Sattler2016PAMI}, which shreds a light on the strengths and weaknesses of each approach. 
Finally, we introduce a new dataset for large-scale indoor localization, consisting of 1,095 high resolution images covering a total area of 5,575$m^2$. Each image is associated with ground truth pose information.
\textcolor{\highlightcolor}{We show that t}his sequence cannot be \textcolor{\highlightcolor}{handled} with SIFT-based methods, as it contains large textureless areas and repetitive structures. \textcolor{\highlightcolor}{In contrast, our approach robustly handles } this scenario and  localize\textcolor{\highlightcolor}{s} images on average within 1\textcolor{\highlightcolor}{.31}m of their ground truth location. 

To summarize, our contribution is three-fold:
	\textcolor{\highlightcolor}{(i) w}e propose a new CNN+LSTM architecture for camera pose regression in indoor and outdoor scenes. \textcolor{\highlightcolor}{Our approach significantly } outperforms \textcolor{\highlightcolor}{previous work on} CNN-based localization \cite{kendall2015posenet,kendall2016bayesianpose}. 
	\textcolor{\highlightcolor}{(ii) w}e provide the first extensive quantitative comparison of CNN-based \textcolor{\highlightcolor}{and} SIFT-based localization methods\textcolor{\highlightcolor}{. We show } that classic SIFT-based methods still outperform all CNN-based methods \textcolor{\highlightcolor}{ by a large margin on existing benchmark datasets}.
	\textcolor{\highlightcolor}{(iii) w}e introduce \textcolor{\highlightcolor}{TUM-LSI\footnote{\textcolor{\highlightcolor}{Dataset available at \url{https://tum-lsi.vision.cs.tum.edu}}},} a new challenging large indoor \textcolor{\highlightcolor}{dataset exhibiting repetitive structures and weakly textured surfaces, and provide} accurate ground truth pose\textcolor{\highlightcolor}{s. We show that CNN-based methods can handle such a challenging scenario while SIFT-based methods fail completely. Thus, we are the first to demonstrate the usefulness of CNN-based methods in practice.} 

\vspace{0.2cm}

\subsection{Related work}
\vspace{0.1cm}

\PAR{Local feature-based localization.}
There are two traditional ways to approach the localization problem. 
Location recognition methods represent a scene by a database of geo-tagged photos. 
Given a query image, they employ image retrieval techniques 
 to identify the database photo most similar to the query \cite{Torii13CVPR,Torii15CVPR,Arandjelovic14ACCV,Sattler16CVPR,Zamir14PAMI}.  
The geo-tag of the retrieved image is often used to approximate the camera pose of the query, even though a more accurate estimate can be obtain by retrieving multiple relevant images \cite{Zamir10ECCV,Zhang06TDPVT,Sattler2017CVPR}. 

More relevant to our approach are structure-based localization techniques that use a 3D model, usually obtained from Structure-from-Motion, to represent a scene. 
They determine the full 6DoF camera pose of a query photo from a set of 2D-3D correspondences established via matching features found in the query against descriptors associated with the 3D points. 
The computational complexity of matching grows with the size of the model. 
Thus, prioritized search approaches \cite{Li10ECCV,Choudhary12ECCV,Sattler2016PAMI} terminate correspondence search as soon as a fixed number of matches has been found. 
Similarly, descriptor matching can be accelerated by using only a subset of all 3D points \cite{Li10ECCV,Cao14CVPR}, which at the same time reduces the memory footprint of the 3D models. 
The latter can also be achieved by quantizing the descriptors \cite{Lynen2015RSS,Sattler15ICCV}.

For more complex scenes, \eg, large-scale urban environments or even large collections of landmark scenes, 2D-3D matches are usually less unique as there often are multiple 3D points with similar local appearance \cite{Li12ECCV}. 
This causes problems for the pose estimation stage as accepting more matches leads to more wrong matches and RANSAC's run-time grows exponentially with the ratio of wrong matches. 
Consequently, Sattler \etal use co-visibility information between 3D points to filter out wrong matches before   pose estimation \cite{Sattler2016PAMI,Sattler15ICCV}.
Similarly, Li \etal use co-visibility information to adapt RANSAC's sampling strategy, enabling them to avoid drawing samples unlikely to lead to a correct pose estimate \cite{Li12ECCV}. 
Assuming that the gravity direction and a rough prior on the camera's height are known, Sv\"{a}rm \etal propose an outlier filtering step whose run-time does not depend on the inlier ratio \cite{Svarm2016PAMI}. 
Zeisl \etal adapt \textcolor{\highlightcolor}{this} approach into a voting scheme, reducing the computational complexity of outlier filtering from $\mathcal{O}(n^2 \log{n})$ \cite{Svarm2016PAMI} to $\mathcal{O}(n)$ for $n$ matches \cite{Zeisl2015ICCV}.

The overall run-time of classical localization approaches depends on the number of features found in a query image, the number of 3D points in the model, and the number of found correspondences and/or the percentage of correct matches. 
In contrast, our approach directly regresses the camera pose from a single feed-forward pass through a network. 
As such, the run-time of our approach only depends on the size of the network used. 

As we will show, SIFT-based methods do not work for our new challenging indoor LSI dataset due to repetitive structures and large textureless regions present in indoor scenes. This further motivates the use alternative approaches based, \eg, on deep learning.

\vspace{0.1cm}

\PAR{Localization utilizing machine learning.}
In order to boost location recognition performance, Gronat \etal and Cao \& Snavely learn linear classifiers on top of a standard bag-of-words \textcolor{\highlightcolor}{vectors} \cite{Cao13CVPR,Gronat13CVPR}. 
They divide the database into distinct places and train classifiers to distinguish between them. 

Donoser \textcolor{\highlightcolor}{\&} Schmalstieg cast feature matching as a classification problem, where the descriptors associated with each 3D model point form a single class \cite{Donoser14CVPR}. 
They employ an ensemble of random ferns to \textcolor{\highlightcolor}{efficiently} compute \textcolor{\highlightcolor}{matches}.

Aubry \etal learn feature descriptors specifically for the task of localizing paintings against 3D scene models \cite{Aubry13TOG}. 

In the context of re-localization for RGB-D images, Guzman-Rivera \etal and Shotton \etal learn random forests that predict a 3D point position for each pixel in an image \cite{shottoncvpr2013,Guzman14CVPR}. 
The resulting 2D-3D matches are then used to estimate the camera pose using RANSAC. 
Rather than predicting point correspondences, Valentin \etal explicitly model the uncertainty of the predicted 3D point positions and use this uncertainty during pose estimation \cite{Valentin15CVPR}, allowing them to localize more images. 
Brachmann \etal adapt the random forest-based approach to not rely on depth measurements during test time 
\cite{brachmanncvpr2016}. 
Still, they require depth data during the training stage as to predict 3D coordinates for each pixel. 
In contrast, 
our approach directly regresses the camera pose from an RGB image, and thus only needs a set of image-pose pairs as input for training.

\PAR{Deep learning.}
CNNs have been successfully applied to most tasks in computer vision since their 
major success in image classification~\cite{krizhevsky12alexnet, simonyan15vgg, he2016resnet} and object detection~\cite{girshick2014rcnn, girshick15fastrcnn, ren2015fasterrcnn}.
One of the major drawbacks of deep learning is its need for large datasets for training. A common approach used for many tasks is that to fine-tune deep architectures 
pre-trained on the seemingly unrelated task of image classification on ImageNet~\cite{russakovsky15imagenet}.
This has been successfuly applied\textcolor{\highlightcolor}{, among others,} to object detection~\cite{girshick15fastrcnn}, object segmentation~\cite{Maninis2016,Kokkinos2016},  semantic segmentation~\cite{hazirbasma16fusenet, noh15deconvnet}, \textcolor{\highlightcolor}{and} depth and normal estimation~\cite{licvpr2015}. 
Similarly, we take pre-trained networks, \eg GoogLeNet~\cite{szedegy2015googlenet}, which can be seen as feature extractors and then fine-tune \textcolor{\highlightcolor}{ them} for the task of camera pose regression.

LSTM \cite{lstm} is a type of Recurrent Neural Network (RNN)~\cite{goler96rnn} designed to accumulate or forget relevant contextual information in its hidden state. It has been successfully applied 
for handwriting recognition \cite{gravesnips2009} \textcolor{\highlightcolor}{and} in natural language processing for machine translation \cite{sutskevernips2014}.
Recently, CNN and LSTM have been combined in the computer vision community to tackle, for example, visual recognition in videos \cite{donahuecvpr2015}.
While most methods in the literature apply LSTM on a temporal sequence, recent works have started to use the memory capabilities of LSTMs to encode contextual information. 
ReNet \cite{renet} replaced convolutions by RNNs sweeping the image vertically and horizontally. 
\cite{VariorECCV2016} uses spatial LSTM for person re-identification, parsing the detection bounding box horizontally and vertically in order to capture spatial dependencies between body parts.  
\cite{byeoncvpr2015} employ\textcolor{\highlightcolor}{ed} the same idea for semantic segmentation and \cite{liangcvpr2016} for semantic object parsing. 
We use LSTMs to better correlate features coming out of the convolutional and FC layers, efficiently reducing feature dimensionality in a structured way that improves pose estimation compared to using dropout on the feature vector to prevent overfitting. \textcolor{\highlightcolor}{A similar approach was simultaneously proposed in \cite{Bell2016CVPR}, where LSTMs are used to obtain contextual information for object recognition.}

End-to-end learning has also been used for localization and location recognition. 
DSAC \cite{Brachmann2017CVPR} proposes a differentiable RANSAC so that a matching function that optimizes pose quality can be learned.
Arandjelovi\'c \etal employ CNNs to learn compact image representations, where each image in a database is represented by a single descriptor \cite{Arandjelovic16CVPR}. 
Weyand \etal cast localization as a classification problem \cite{weyandeccv2016}. 
They adaptively subdivide the earth's surface in a set of tiles, where a finer quantization is used for regions exhibiting more images. 
The CNN then learns to predict the corresponding tile for a given image, thus providing the approximate position from which a photo was taken. 
Focusing on accurate 6DoF camera pose estimation, the PoseNet method by Kendall \etal uses CNNs to model pose estimation as a regression problem \cite{kendall2015posenet}. 
An extension of the approach repeatedly evaluates the CNN with a fraction of its neurons randomly disabled, resulting in multiple different pose \textcolor{\highlightcolor}{estimates} that can be used to predict pose uncertainty \cite{kendall2016bayesianpose}. 
One drawback of the PoseNet approach is its relative inaccuracy \cite{brachmanncvpr2016}. 
In this paper, we show how a CNN+LSTM architecture is able to produce significantly more accurate camera poses \textcolor{\highlightcolor}{compared to PoseNet}.   

\begin{figure*}[t]
	\centering
  \includegraphics[width=0.9\linewidth]{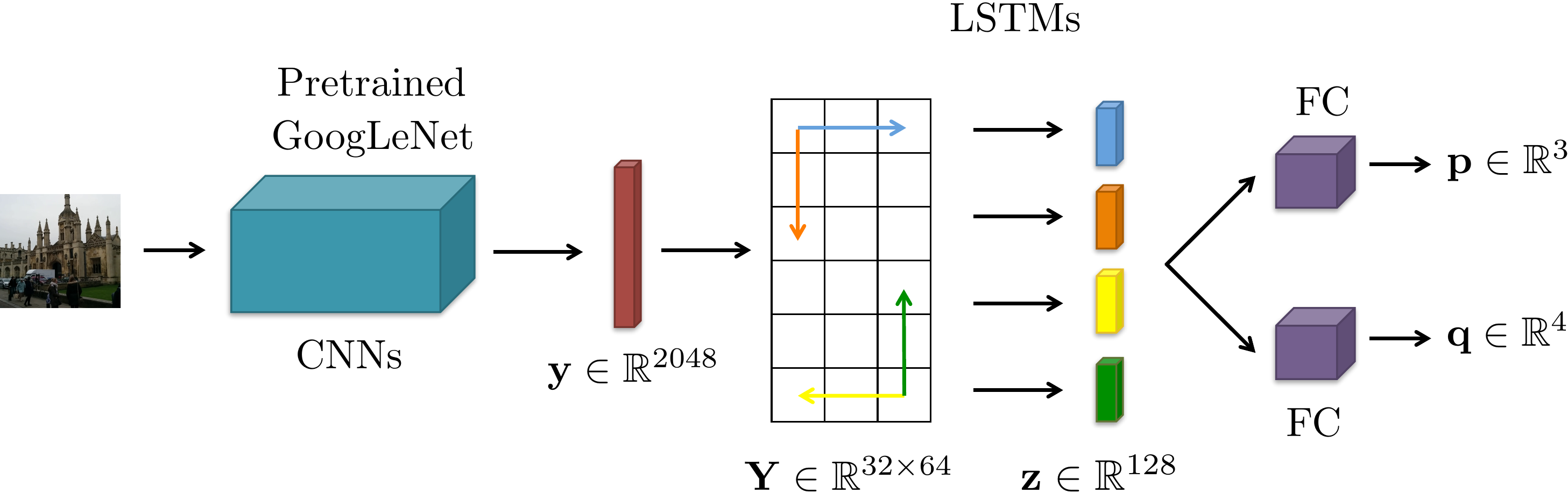}
	\caption{Architecture of the proposed pose regression LSTM network.}%
	\vspace{-6pt}
	\label{fig:architecture}
\end{figure*}

\section{Deep camera pose regression}

In this section, we develop our framework for learning to regress camera poses directly from images. Our goal is to train a CNN+LSTM network to learn a mapping from an image to a pose, 
$I \stackrel{f}{\rightarrow} \mathbf{P}$, where $f(\cdot)$ is the neural network.
Each pose $\mathbf{P}=[\mathbf{p},\mathbf{q}]$ is represented by its 3D camera position $\mathbf{p} \in \mathbb{R}^3 $ and a quaternion $\mathbf{q} \in \mathbb{R}^4 $ for its orientation.

Given a dataset composed of training images $I_i$ and their corresponding 3D ground truth poses $\mathbf{P}_i$, we train the network using Adam \cite{adam} with the \textcolor{\highlightcolor}{same} loss function \textcolor{\highlightcolor}{already used in  \cite{kendall2015posenet}}:

\begin{equation}
	L_i =\norm{\mathbf{p}_i - \hat{\mathbf{p}}_i}_{2} + \beta \cdot \norm{ \mathbf{q}_i- \frac{ \hat{\mathbf{q}}_i}{\norm{ \hat{\mathbf{q}}_i}}}_{2} \enspace ,
	\label{eq:loss}
\end{equation}
where $(\mathbf{p},\mathbf{q})$ and $(\hat{\mathbf{p}},\hat{\mathbf{q}})$ are ground truth and estimated position-orientation pairs, respectively. 
We represent the orientation with quaternions and thus normalize the predicted orientation $\hat{\mathbf{q}_i}$  to unit length.
~$\beta$ determines the relative weight of \textcolor{\highlightcolor}{the} orientation error \textcolor{\highlightcolor}{wrt.} to \textcolor{\highlightcolor}{the} positional error and is in general bigger for outdoor scenes, as errors tend to be relatively larger \textcolor{\highlightcolor}{\cite{kendall2015posenet}}. All hyperparameters used for the experiments are
detailed in Section \ref{experiments}.

\subsection{CNN architecture: feature extraction}

Training a neural network from scratch for the task of pose regression would be impractical for several reasons: (i) we would need a really large training set, (ii) compared to classification problems, 
where each output label is covered by at least one training sample, the output in regression is continuous and infinite. 
Therefore, we leverage a pre-trained classification network, namely GoogLeNet~\cite{szedegy2015googlenet}, and modify it in a similar fashion as \textcolor{\highlightcolor}{in}~\cite{kendall2015posenet}.  
At the end of the convolutional layers average pooling is performed, followed by a fully connected layer which outputs a 2048 dimensional vector \textcolor{\highlightcolor}{(\cf}  Figure~\ref{fig:architecture}\textcolor{\highlightcolor}{)}. 
This can be seen as a feature vector that represents the image to be localized. 
This architecture is 
used \textcolor{\highlightcolor}{in}~\cite{kendall2015posenet} to predict camera poses by using yet another fully connected regression layer at the end that outputs the 7-dimensional pose and orientation vector (the quaternion vector is normalized to unit length at test time).

\subsection{Structured feature correlation with LSTMs}

After the convolutional layers of GoogleNet, an average pooling layer gathers the information of each feature channel for the entire image. Following PoseNet \cite{kendall2015posenet}, we use a fully connected (FC) layer after pooling to learn the correlation among features.
As we can see from the PoseNet \cite{kendall2015posenet} results \textcolor{\highlightcolor}{shown in Section~\ref{experiments}}, regressing the pose after the high dimensional output of a fully connected (FC) layer is not optimal. Intuitively, \textcolor{\highlightcolor}{the dimensionality of the 2048D embedding of the image through the FC layer is typically relatively large compared to the amount of available training data. As a result, the linear pose regressor has many degrees of freedom and it is likely that overfitting leads to inaccurate predictions for test images dissimilar to the training images.}
One could directly reduce \textcolor{\highlightcolor}{the} dimension\textcolor{\highlightcolor}{s} of the FC, but we empirically found that dimensionality reduction performed by \textcolor{\highlightcolor}{a} network with \textcolor{\highlightcolor}{LSTM} memory blocks is more \textcolor{\highlightcolor}{effective}. 
\textcolor{\highlightcolor}{Compared to applying dropout within PoseNet to avoid overfitting~\cite{kendall2016bayesianpose}, our approach consistently estimates  more accurate positions, which justifies our use of LSTMs.}


Even though Long Short-Term Memory (LTSM) units have been typically applied to temporal sequences, recent works \cite{Bell2016CVPR,renet,VariorECCV2016,byeoncvpr2015,liangcvpr2016} have used the memory capabilities of LSTMs in image space. 
In our case, we treat the output 2048 feature vector as our sequence.
We propose to insert four LSTM units after the FC, which have the function of reducing the dimensionality of the feature vector in a structured way. The memory units \textcolor{\highlightcolor}{identify} 
the most useful feature correlations for the task of pose estimation.

\PAR{Reshaping the input vector.} In practice, inputting the 2048-D vector directly to the LSTM did not show good results. Intuitively, this is because even though the memory unit of the LSTM is capable of remembering distant features, a 2048 length vector is too long for LSTM to correlate from the first to the last feature.
We thereby propose to reshape the vector to a $32 \times 64$ matrix and to apply four LSTMs in the up, down, left and right directions as depicted in Figure~\ref{fig:architecture}. These four outputs are then concatenated and passed to the fully connected pose prediction layers.
This imitates the function of structured dimensionality reduction which greatly improves pose estimation \textcolor{\highlightcolor}{accuracy}.

\begin{figure*}[ht]
\centering
\begin{subfigure}{0.165\linewidth}
\centering
\includegraphics[width=0.98\textwidth]{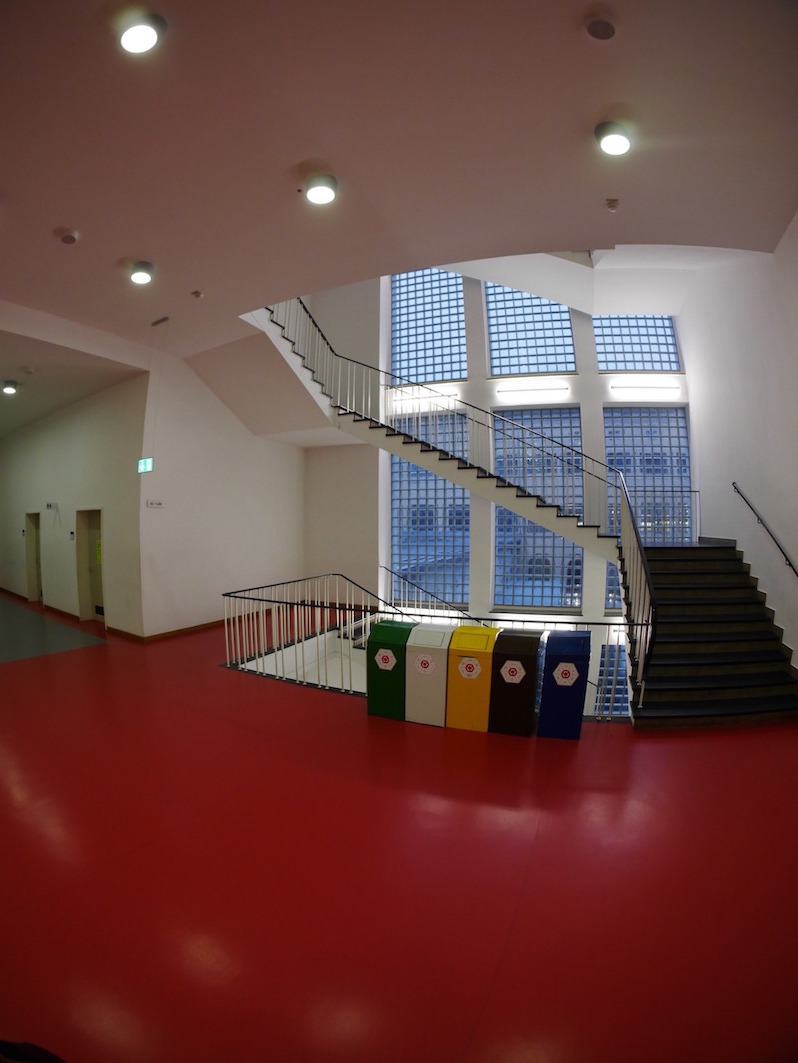}
\end{subfigure}%
\begin{subfigure}{0.165\linewidth}
\centering
\includegraphics[width=0.98\textwidth]{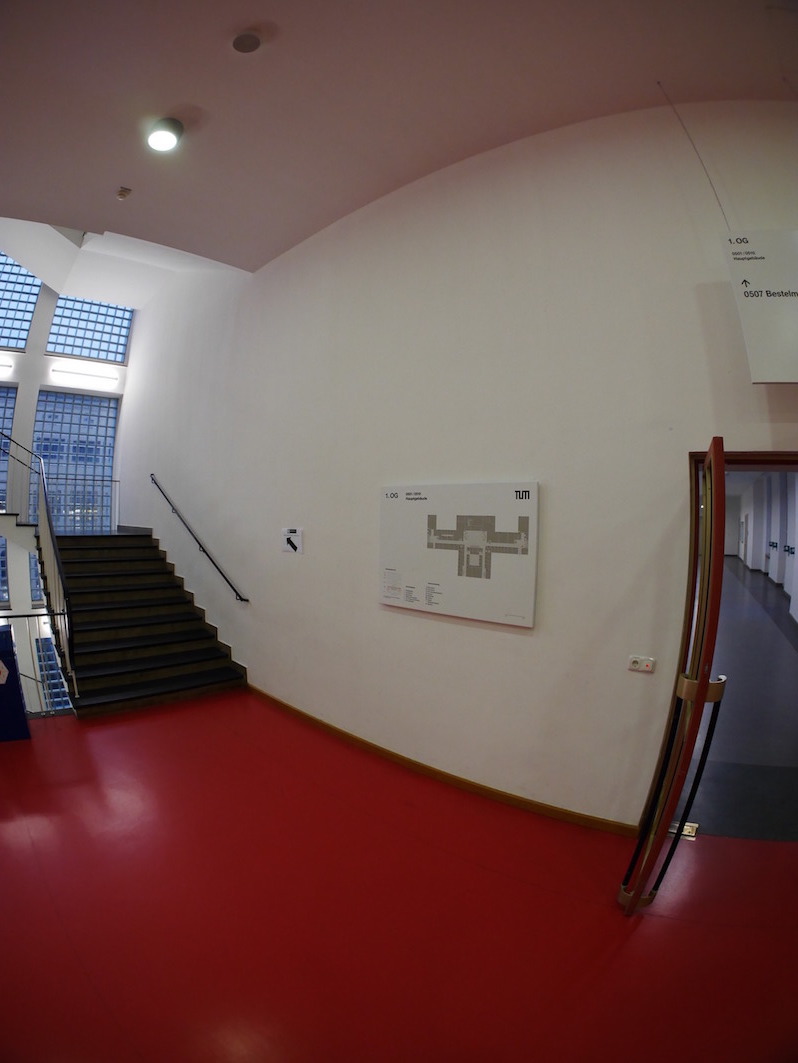}
\end{subfigure}%
\begin{subfigure}{0.165\linewidth}
\centering
\includegraphics[width=0.98\textwidth]{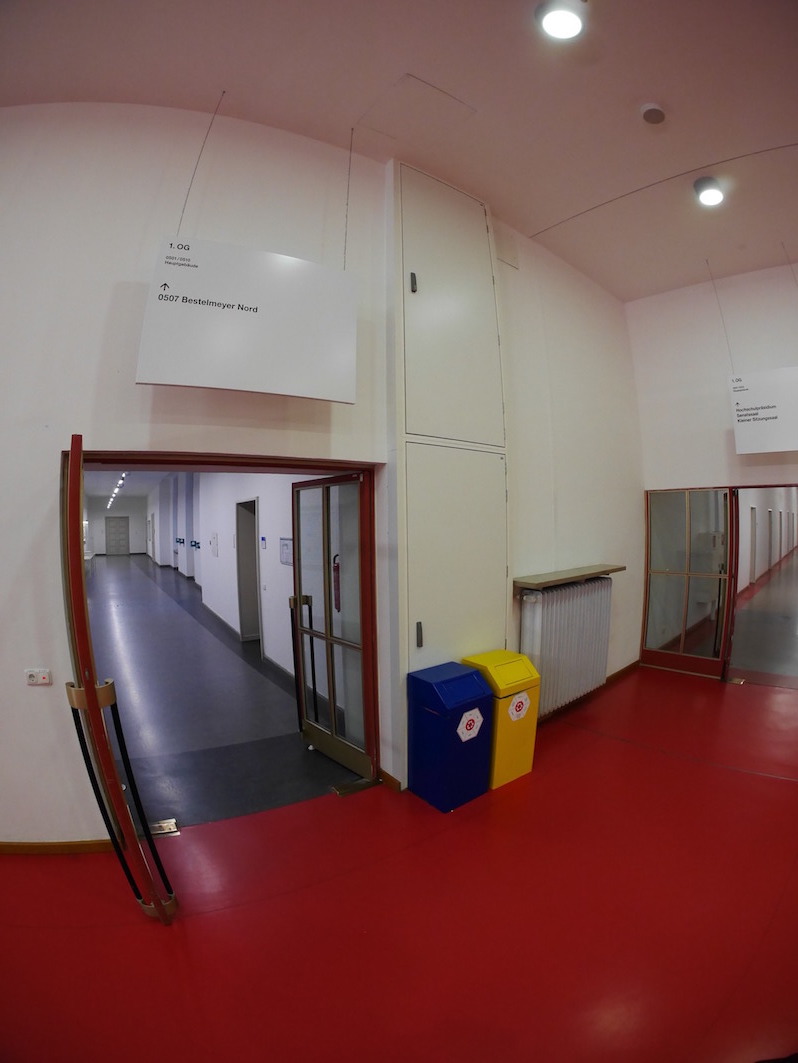}
\end{subfigure}%
\begin{subfigure}{0.165\linewidth}
\centering
\includegraphics[width=0.98\textwidth]{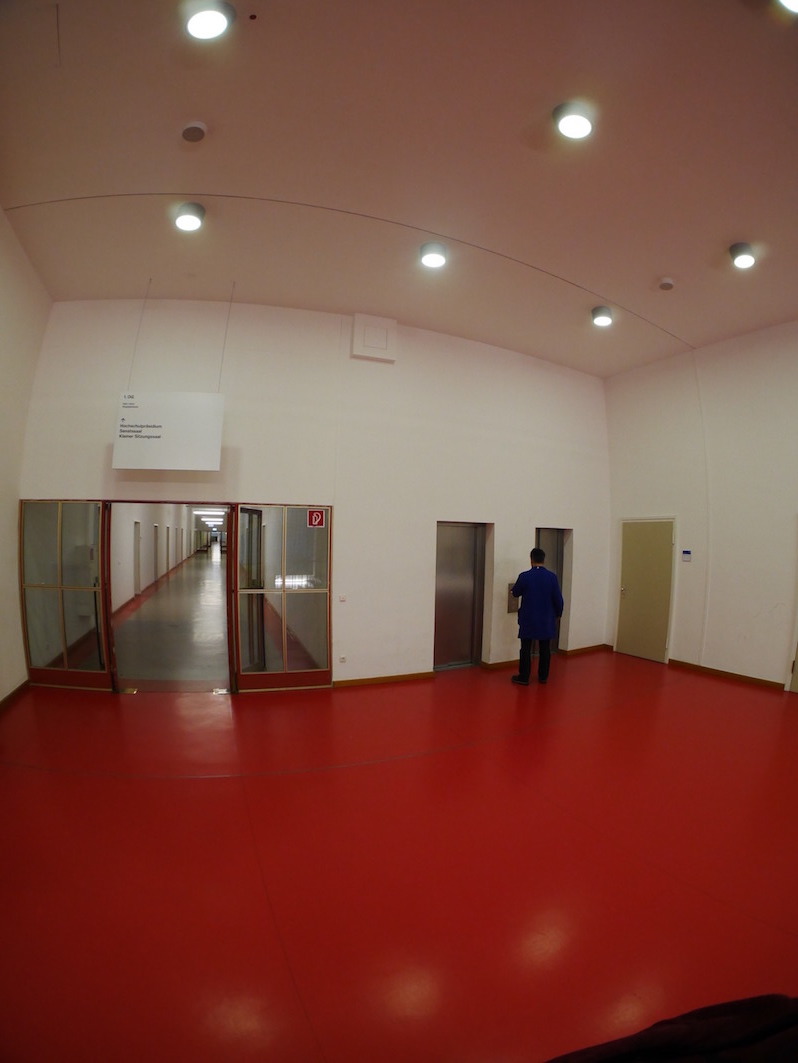}
\end{subfigure}%
\begin{subfigure}{0.165\linewidth}
\centering
\includegraphics[width=0.98\textwidth]{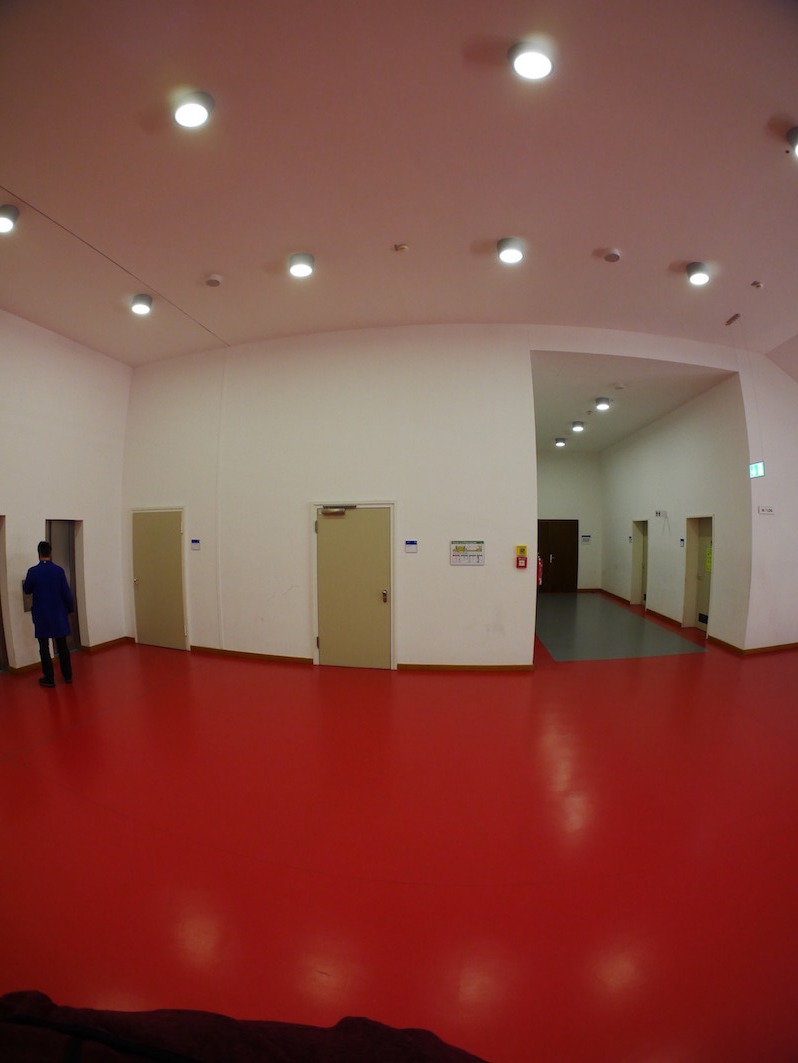}
\end{subfigure}%
\begin{subfigure}{0.165\linewidth}
\centering
\includegraphics[width=0.98\textwidth]{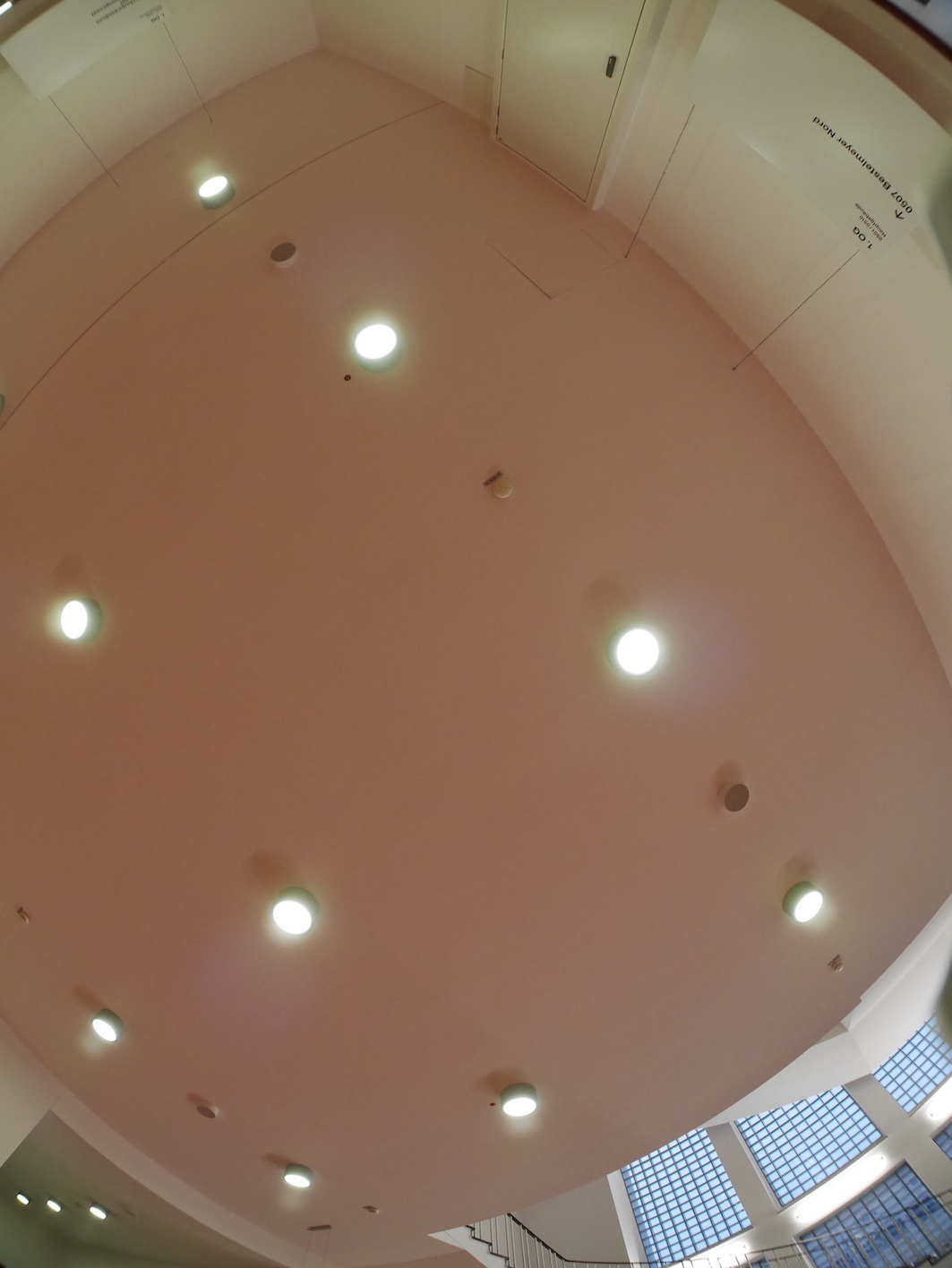}
\end{subfigure}
\caption{Example images from our \textcolor{\highlightcolor}{TUM-LSI dataset}. At each capture-location, we provide a set of six high-resolution wide-angle pictures, taken in five different horizontal directions and one pointing up.}
\label{fig:lsi-example-images}
\end{figure*}

\section{A large-scale indoor localization \textcolor{\highlightcolor}{dataset}}
Machine learning and in particular deep learning are inherently data-intensive endeavors. Specifically, supervised learning requires not only data but also associated ground truth labelling. 
For some tasks such as image classification~\cite{russakovsky15imagenet} or outdoor image-based localization~\cite{kendall2015posenet} large training and testing datasets have already been made available to the community. For indoor scenes, only small datasets covering a spatial extent the size of a room~\cite{shottoncvpr2013} are currently available.


We introduce the \emph{TU Munich \textcolor{\highlightcolor}{Large-Scale Indoor}} \textcolor{\highlightcolor}{(TUM-LSI) dataset} covering an area that is two orders of magnitude larger than \textcolor{\highlightcolor}{the typically used } 7Scenes \textcolor{\highlightcolor}{dataset}~\cite{shottoncvpr2013}.
It comprises 1,095 high-resolution images ($4592\times3448$ \textcolor{\highlightcolor}{pixels}) with geo-referenced pose information for each image. The dataset spans a whole building floor with a total area of 5,575 \textcolor{\highlightcolor}{$m^2$}. 
Image locations are spaced roughly one meter apart, and at \textcolor{\highlightcolor}{location} each we provide a set of six wide-angle pictures, taken in five different horizontal directions (full $360^{\circ}$) and one pointing up (see Figure~\ref{fig:lsi-example-images}).  
\textcolor{\highlightcolor}{Our new dataset is} very challenging \textcolor{\highlightcolor}{due to} repeated structural elements with nearly identical appearance, e.g. two \textcolor{\highlightcolor}{nearly identical} stair cases, \textcolor{\highlightcolor}{that create} global ambiguities. Notice that such global ambiguities often only appear at larger scale and are thus missing from the 7Scenes dataset. In addition, there is a general lack of well-textured regions, whereas 7Scenes mostly depict highly textured areas. Both problems make this \textcolor{\highlightcolor}{dataset} challenging to approaches that only considers (relatively) small image patches. 

In order to generate ground truth pose information for each image, we captured the data using the \emph{NavVis M3}\footnote{www.navvis.com} indoor mapping platform. This mobile system is equipped with six \emph{Panasonic} 16-Megapixel system cameras and three \emph{Hokuyo} laser range finders. Employing SLAM, the platform is able to reconstruct the full trajectory with sub-centimeter accuracy. 

\section{Experimental results}
\label{experiments}

We present results on several datasets, proving the efficacy of our method in outdoor scenes like Cambridge \cite{kendall2015posenet} and small-scale indoor scenes such as 7Scenes \cite{shottoncvpr2013}. 
The two datasets are very different from each other: 7Scenes has a very high number of images in a very small spatial extent, hence, it is more suited for applications such as Augmented Reality, while Cambridge
Landmarks has sparser coverage and larger spatial extent, the perfect scenario for image-based localization. 
In the experiments, we show \textcolor{\highlightcolor}{that} our method can be applied to both scenarios \textcolor{\highlightcolor}{and} deliver\textcolor{\highlightcolor}{s} competitive results.
We provide comparisons to previous CNN-based approaches, as well as a state-of-the-art SIFT-based localization method~\cite{Sattler2016PAMI}. 
Furthermore, we provide results for our new \textcolor{\highlightcolor}{TUM-LSI dataset}. SIFT-based methods fail on \textcolor{\highlightcolor}{TUM-LSI} due to textureless surfaces and repetitive structures, while our method is able to localize images with an average accuracy of 1.3\textcolor{\highlightcolor}{1}m for an area of 5,575 \textcolor{\highlightcolor}{$m^2$}. 

\PAR{Experimental setup.}
We initialize the GoogLeNet part of the network with the Places \cite{Places} weights and randomly initialize the remaining weights.
All networks take images of size $224\times224$ pixel as input. We use random crops during training and central crops during testing. A mean image is computed separately for each training sequence and is subtracted from all images.
All experiments are performed on an NVIDIA Titan X using TensorFlow with Adam~\cite{adam} for optimization. Random shuffling is performed for each batch, and regularization is only applied to weights, not biases. 
For all sequences we use the following hyperparameters: batch size 75, regularization $\lambda=2^{-4}$, auxiliary loss weights $\gamma=0.3$, dropout probability 0.5, and the parameters for Adam: $\epsilon=1$, $\beta_1=0.9$ and $\beta_2=0.999$.
The $\beta$ of Eq.~\ref{eq:loss} balances the orientation and positional penalties. To ensure a fair comparison, for Cambridge Landmarks and 7Scenes, we take the same values as PoseNet~\cite{kendall2015posenet}: for the indoor scenes $\beta$ is between
120 to 750 and outdoor scenes between 250 to 2000.
For \textcolor{\highlightcolor}{TUM-LSI}, 
we set $\beta=1000$.

\begin{table*}[t]
  \newcommand*{\poseerr}[2]{\SI{#1}{\m}, \SI{#2}{\degree}}
  \centering
  \caption{Median localization results of several RGB-only methods on Cambridge Landmarks~\cite{kendall2015posenet} and 7Scenes~\cite{shottoncvpr2013}. }
  \small
\begin{tabular}{@{} p{2.3cm} M{1.2cm} M{2.4cm}  M{2.4cm}   M{1.8cm}   M{1.8cm}  M{3cm}  @{}} \toprule
  Scene & Area or Volume & Active Search (w/o) \cite{Sattler2016PAMI} & Active Search (w/) \cite{Sattler2016PAMI} & PoseNet \cite{kendall2015posenet} & Bayesian PoseNet\cite{kendall2016bayesianpose} & Proposed + Improvement(pos,ori) \\ \midrule
  King's College   & 5600 $m^2$ & \poseerr{0.42}{0.55} (0) & \poseerr{0.57}{0.70} (0) & \poseerr{1.92}{5.40} & \poseerr{1.74}{4.06} & \poseerr{0.99}{3.65} (48,32) \\
  Old Hospital     & 2000 $m^2$ &  \poseerr{0.44}{1.01} (2)  & \poseerr{0.52}{1.12} (2) & \poseerr{2.31}{5.38} & \poseerr{2.57}{5.14} & \poseerr{1.51}{4.29} (35,20)\\
  Shop Fa\c{c}ade  & 875 $m^2$ & \poseerr{0.12}{0.40} (0) & \poseerr{0.12}{0.41} (0) & \poseerr{1.46}{8.08} & \poseerr{1.25}{7.54} & \poseerr{1.18}{7.44} (19,8)\\
  St Mary's Church & 4800 $m^2$ & \poseerr{0.19}{0.54} (0) & \poseerr{0.22}{0.62} (0) & \poseerr{2.65}{8.48} & \poseerr{2.11}{8.38} & \poseerr{1.52}{6.68} (43,21) \\ \midrule
  Average All          &  -- & -- & -- & \poseerr{2.08}{6.83} & \poseerr{1.92}{6.28} & \poseerr{1.30}{5.52} (37,19)\\ 
  Average by \cite{Sattler2016PAMI}  & -- & \poseerr{0.29}{0.63} & \poseerr{0.36}{0.71} & -- & -- &  \poseerr{1.37}{5.52}  \\ 
  \midrule
  \\ \midrule
  Chess            & 6 $m^3$ & \poseerr{0.04}{1.96} (0) & \poseerr{0.04}{2.02} (0)& \poseerr{0.32}{8.12} & \poseerr{0.37}{7.24} & \poseerr{0.24}{5.77} (25,29) \\
  Fire             & 2.5 $m^3$ & \poseerr{0.03}{1.53} (1) & \poseerr{0.03}{1.50} (1) & \poseerr{0.47}{14.4} & \poseerr{0.43}{13.7} & \poseerr{0.34}{11.9} (28,17) \\
  Heads            & 1 $m^3$ & \poseerr{0.02}{1.45} (1) & \poseerr{0.02}{1.50} (1) & \poseerr{0.29}{12.0} & \poseerr{0.31}{12.0} & \poseerr{0.21}{13.7} (27,-14)\\
  Office           & 7.5 $m^3$ & \poseerr{0.09}{3.61} (34) & \poseerr{0.10}{3.80} (34) & \poseerr{0.48}{7.68} & \poseerr{0.48}{8.04} & \poseerr{0.30}{8.08} (37,-5) \\
  Pumpkin          & 5 $m^3$ & \poseerr{0.08}{3.10} (71) & \poseerr{0.09}{3.21} (68) & \poseerr{0.47}{8.42} & \poseerr{0.61}{7.08} & \poseerr{0.33}{7.00} (30,17) \\
  Red Kitchen      & 18 $m^3$ & \poseerr{0.07}{3.37} (0) & \poseerr{0.07}{3.52} (0) & \poseerr{0.59}{8.64} & \poseerr{0.58}{7.54} & \poseerr{0.37}{8.83} (37,-2) \\
  Stairs           & 7.5 $m^3$ & \poseerr{0.03}{2.22} (3)   & \poseerr{0.04}{2.22} (0)        & \poseerr{0.47}{13.8} & \poseerr{0.48}{13.1} & \poseerr{0.40}{13.7} (15,0.7) \\ \midrule
  Average  All        & -- & -- & -- & \poseerr{0.44}{10.4} & \poseerr{0.47}{9.81} & \poseerr{0.31}{9.85} (29,5)\\
  Average  by \cite{Sattler2016PAMI}         & -- & \poseerr{0.05}{2.46} & \poseerr{0.06}{2.54} & -- & -- &  \poseerr{0.30}{9.15}\\\bottomrule
\end{tabular}

\label{tab:cambridge7scenes}
  \vspace{-0.3cm}
\end{table*}

\PAR{Comparison with state-of-the-art.}
We compare results to two CNN-based approaches: PoseNet~\cite{kendall2015posenet} and Bayesian PoseNet~\cite{kendall2016bayesianpose}.
On Cambridge Landmarks and 7Scenes, results for the two PoseNet variants~\cite{kendall2015posenet,kendall2016bayesianpose} were taken directly from the author's publication~\cite{kendall2016bayesianpose}. For the new \textcolor{\highlightcolor}{TUM-LSI dataset}, their model was fine-tuned with the training images. The hyperparameters used are the same as for our method, except for \textcolor{\highlightcolor}{the} Adam parameter $\epsilon=0.1$, which showed better convergence.

To the best of our knowledge, CNN-based approaches have not been quantitatively compared to SIFT-based localization approaches. We feel this comparison is extremely important to know how deep learning can make an impact in image-based localization, and what challenges are there to overcome. We therefore present results of a state-of-the-art SIFT-based method, namely Active Search~\cite{Sattler2016PAMI}. 

Active Search estimates the camera poses wrt. a SfM model, where each 3D point is associated with SIFT descriptors extracted from the training images. 
Since none of the datasets provides both an SfM model and the SIFT descriptors, we constructed such models from scratch using VisualSFM \cite{Wu11CVPR,Wu133DV} and COLMAP \cite{schoenberger2016sfm} and registered them against the ground truth poses of the training images. \textcolor{\highlightcolor}{Thus, the camera poses reported for Active Search contain both the errors made by Active Search and  the reconstruction and registration processes.} 
The models used for localization do not contain any contribution from the testing images. 

Active Search uses a visual vocabulary to accelerate descriptor matching. 
We trained a vocabulary containing 10k words from training images of the Cambridge dataset and a vocabulary containing 1k words from training images of the smaller 7Scenes dataset. 
Active Search uses these vocabularies for prioritized search for efficient localization, where matching is terminated once a fixed number of correspondences has been found. 
We report results both with (w/) and without (w/o) prioritization. 
In the latter case, we simply do not terminate matching early but try to find as many correspondences as possible. 
For querying with Active Search, we use calibrated cameras with a known focal length, obtained from the SfM reconstructions, but ignore radial distortion. 
As such, camera poses are estimated using a 3-point-pose solver \cite{Kneip2011CVPR} inside a RANSAC loop \cite{Fischler81CACM}. 
%
Poses estimated from only few matches are usually rather inaccurate. 
Following common practice \cite{Li10ECCV,Li12ECCV}, Active Search only considers a testing image as successfully localized if its pose was estimated from at least 12 inliers.



\begin{figure}[b]
	\vspace{-0.2cm}
	\centering
	\begin{subfigure}{0.3\linewidth}
	\centering
	\includegraphics[width=\textwidth]{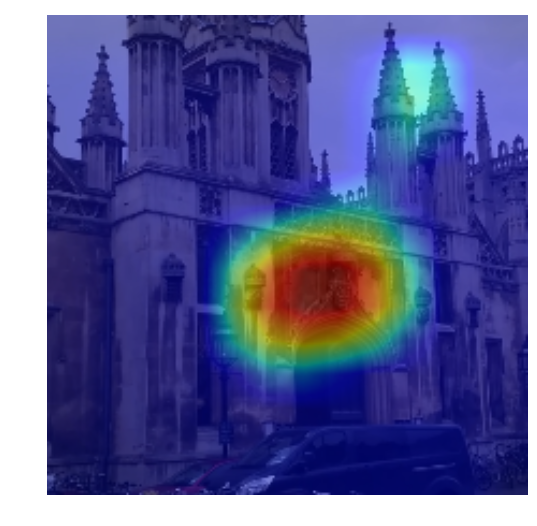}
	\end{subfigure}%
	\begin{subfigure}{0.3\linewidth}
	\centering
	\includegraphics[width=\textwidth]{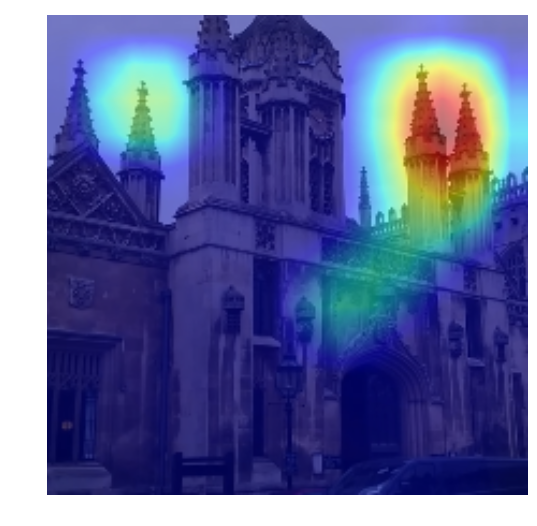}
	\end{subfigure}%
	\begin{subfigure}{0.3\linewidth}
	\centering
	\includegraphics[width=\textwidth]{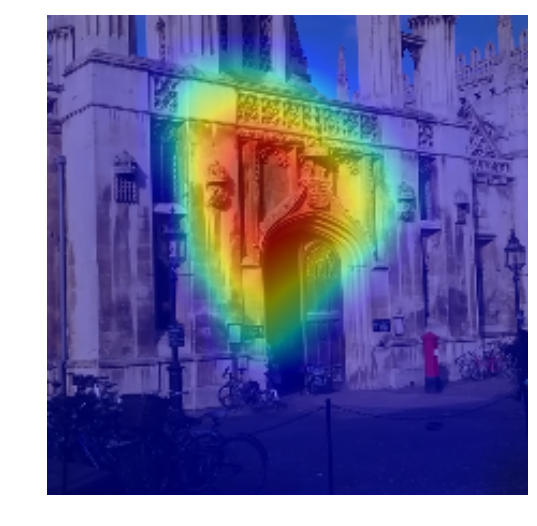}
	\end{subfigure}%
	\vspace{-6pt}%
	\caption[Class activation map for King's College]{The class activation map is overlaid on an input image from King's College as a heat map. Red areas indicate parts of the image the network considers important for pose regression. The visualization shows how the network focuses on distinctive building elements.}
	\vspace{-0.4cm}
	\label{fig:activationmap}
\end{figure}

\subsection{Large-scale outdoor localization}

We present results for outdoor image-based localization on the publicly available Cambridge Landmarks dataset\cite{kendall2015posenet} in Table~\ref{tab:cambridge7scenes}.
We report results for Active Search only for images with at least 12 inliers and give the number of images where localization fails in parenthesis. 
In order to compare the methods fairly, we provide the average accuracy for all images (Average All), and also the average accuracy for only those images that ~\cite{Sattler2016PAMI} was able to localize (Average by~\cite{Sattler2016PAMI}).
Note, that we do not report results on the Street dataset of Cambridge Landmarks. It is a unique sequence because the training database consists of four distinct video sequences, 
each filmed in a different compass direction. This results in training images at similar positions, but with very different orientations.
Even with the hyperparameters set by the author of~\cite{kendall2015posenet}, training did not converge for any of the implemented methods. 

\begin{figure*}[htpb]
	\centering
	\begin{subfigure}{0.33\linewidth}
	\centering
	\includegraphics[height=3cm]{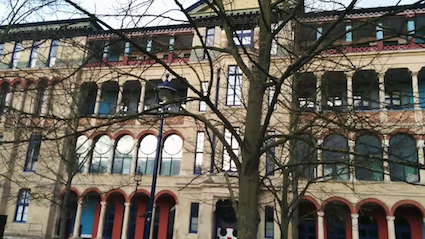}
	\caption{Original image}
	\end{subfigure}
	\begin{subfigure}{0.33\linewidth}
	\centering
	\includegraphics[height=3cm]{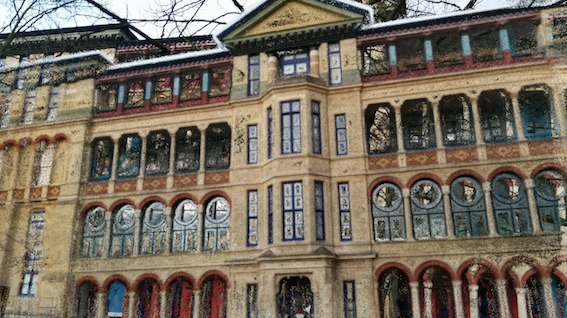}
	\caption{Posenet result~\cite{kendall2015posenet}}
	\end{subfigure}
	\begin{subfigure}{0.33\linewidth}
	\centering
	\includegraphics[height=3cm]{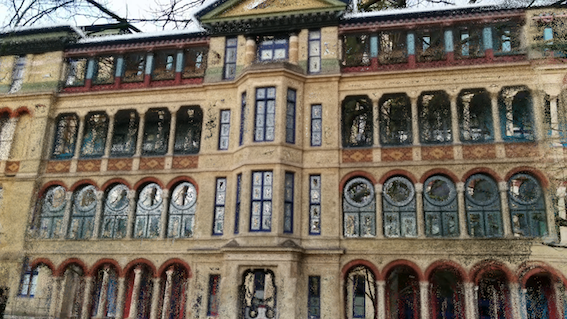}
	\caption{Our result}
	\end{subfigure}%
	\vspace{-6pt}%
	\caption{Example\textcolor{\highlightcolor}{s from} the Old Hospital sequence. \textcolor{\highlightcolor}{The } 3D \textcolor{\highlightcolor}{scene} model is \textcolor{\highlightcolor}{projected into the image using the poses estimated by } (b) PoseNet~\cite{kendall2015posenet}  and (c) our \textcolor{\highlightcolor}{method}. Active Search~\cite{Sattler2016PAMI} \textcolor{\highlightcolor}{did} not localize the image due to the occlusion \textcolor{\highlightcolor}{caused} by the tree. Note the inaccuracy of PoseNet compared to the proposed method (check the top of the building for alignment).}
		\vspace{-0.2cm}
	\label{fig:old_hospital}
\end{figure*}

In parenthesis next to our results, we show the rounded percentage improvement wrt. PoseNet in both position and orientation, separated by a comma.
As we can see, the proposed method on average reduces positional error by 37.5\% wrt. PoseNet and the orientation error by 19\%. 
For example, in King's College the positional error is reduced by more than 40\%, going from 1.92m \textcolor{\highlightcolor}{for} Posenet to 0.99m for our method. 
This shows that the proposed LSTM-based structured output is efficient in encoding feature correlations, leading to large improvements in localization performance.
It is important to note that none of the CNN-based methods is able to match the precision of \textcolor{\highlightcolor}{Active Search}~\cite{Sattler2016PAMI}, especially when computing the orientation of the camera. 
Since~\cite{Sattler2016PAMI} requires 12 inliers to consider an image as localized, it is able to reject inaccurate poses. 
In contrast, our method always provides a localization result, even if it is sometimes less accurate. Depending on the application, one behavior or the other might be more desirable. 
As an example, we show in Figure~\ref{fig:old_hospital} an image from Old Hospital, where a tree is occluding part of the building. In this case, \cite{Sattler2016PAMI} is not able to localize the image, while our method still produces a reasonably accurate pose. 
This phenomenon becomes more important in indoor scenes, where Active Search is unable to localize a substantially larger number of images\textcolor{\highlightcolor}{, mostly due to motion blur}.

\begin{figure*}[htpb]
	\centering
	\begin{subfigure}{0.33\linewidth}
	\centering
	\includegraphics[width=5cm]{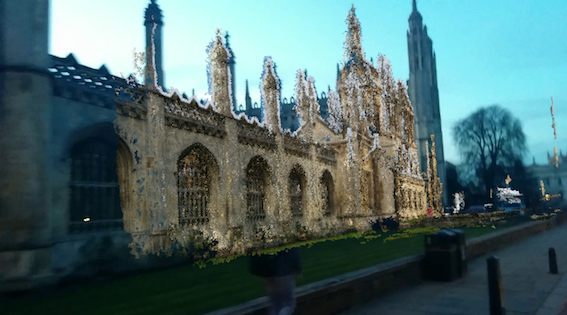}
	\end{subfigure}
	\begin{subfigure}{0.33\linewidth}
	\centering
	\includegraphics[width=5cm]{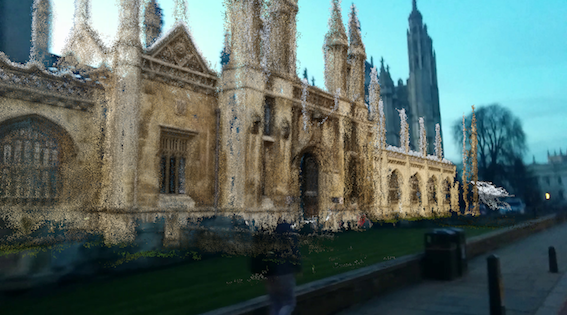}
	\end{subfigure}
	\begin{subfigure}{0.33\linewidth}
	\centering
	\includegraphics[width=5cm]{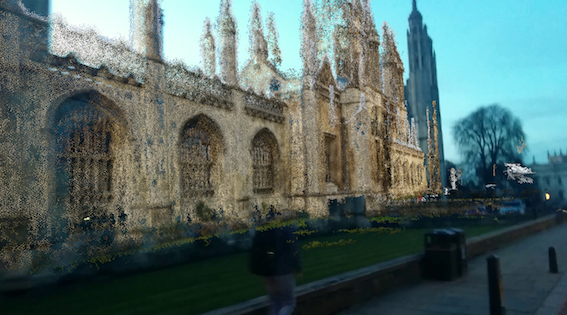}
	\end{subfigure}%
	\centering \\ \vspace{0.2cm}
	\begin{subfigure}{0.33\linewidth}
	\centering 
	\includegraphics[width=5cm]{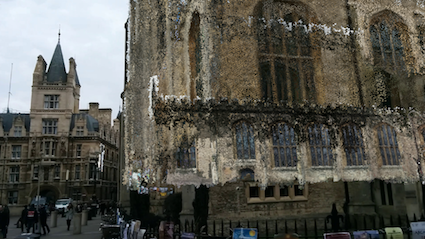}
	\caption{Active Search (w/)~\cite{Sattler2016PAMI} result}
	\end{subfigure}
	\begin{subfigure}{0.33\linewidth}
	\centering
	\includegraphics[width=5cm]{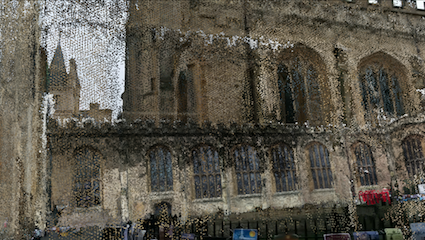}
	\caption{Posenet result~\cite{kendall2015posenet}}
	\end{subfigure}
	\begin{subfigure}{0.33\linewidth}
	\centering
	\includegraphics[width=5cm]{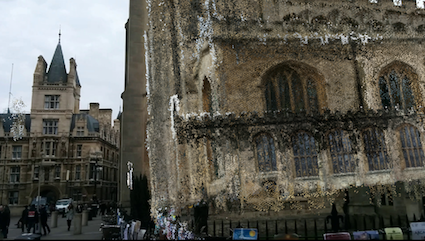}
	\caption{Our result}
	\end{subfigure}%
	\vspace{-6pt}%
	\caption{Examples of localization results on King's College for Active Search~\cite{Sattler2016PAMI}, PoseNet~\cite{kendall2015posenet}, and the proposed method.}
	\label{fig:visualresults}
		\vspace{-0.2cm}
\end{figure*}

Interestingly, for our method, the average for all images is lower than the average only for the images that ~\cite{Sattler2016PAMI} can also localize. This means that for images where~\cite{Sattler2016PAMI} cannot return a pose, our method actually provides a very accurate result. This ``complementary" behavior between SIFT- and CNN-based methods could be exploitable in future research.
Overall, our method shows a strong performance in outdoor image-based localization, as seen in Figure \ref{fig:visualresults}, where \textcolor{\highlightcolor}{PoseNet}~\cite{kendall2015posenet} provides less accurate poses.
In order to better understand how the network localizes an image, \textcolor{\highlightcolor}{Figure~\ref{fig:activationmap} plots} the class activation maps for the King's College sequence. Notice how strong activations cluster around distinctive building elements, \textcolor{\highlightcolor}{\eg, the towers and entrance}.

\subsection{Small-scale indoor localization}

In this section, we focus on localization on small indoor spaces, for which we use the publicly available 7Scenes dataset~\cite{shottoncvpr2013}. 
Results at the bottom of Table~\ref{tab:cambridge7scenes} show that we also outperform previous CNN-based PoseNet by 29\% in positional error and 5.3\% orientation error. 
For example on, Pumpkin we achieve a positional error reduction from the 0.61m for Posenet to 0.33m for our method. 
\textcolor{\highlightcolor}{As for the Cambridge Landmarks dataset, we observe that our approach consistently outperforms Bayesian PoseNet~\cite{kendall2016bayesianpose}, which uses dropout to limit overfitting during training. These experimental results validate our strategy of using LSTMs for structured dimensionality reduction in an effort to avoid overfitting.}

There are two methods that use RGB-D data and achieve a lower error but still higher than Active Search: \cite{brachmanncvpr2016} achieves 0.06m positional error and 2.7\degree orientation error, while \cite{shottoncvpr2013} scores 0.08m and 1.60\degree. 
Note, that these methods require RGB-D data for training \cite{brachmanncvpr2016} and/or testing \cite{shottoncvpr2013}. 
It is unclear though how well such methods would work in outdoor scenarios with stereo data. 
In theory, multi-view stereo methods could be used to obtain the required depth maps for outdoor scenes.  
However, such depth maps are usually substantially more noisy and contain significantly more outliers compared to the data obtained with RGB-D sensors. 
In addition, the accuracy of the depth maps decreases quadratically with the distance to the scene, which is usually much larger for outdoor scenes than for indoor scenes. 
In \cite{Valentin15CVPR}, authors report that 63.4\% of all test images for Stairs can be localized with a position error less than 5cm and an orientation error smaller than 5$^\circ$. 
With and without prioritization, Active Search localizes 77.8\% and 80.2\%, respectively, within these error bounds. 
Unfortunately, median registration errors of 2-5cm observed when registering the other SfM models against the 7Scene datasets prevent us from a more detailed comparison. 
%

As we can see in Table~\ref{tab:cambridge7scenes}, if an image can be localized, we notice that Active Search performs better than CNN-based approaches. However, we note that for Office and Pumpkin the number of images not localized is fairly large, 34 and 71, respectively.
We provide the average accuracy for all images (Average All), and also the average accuracy for only those images that ~\cite{Sattler2016PAMI} was able to localize (Average by~\cite{Sattler2016PAMI}). Note that for our method, the two averages are extremely similar, \textcolor{\highlightcolor}{\ie,} we are able to localize those images with the same accuracy as all the rest, showing robustness wrt. \textcolor{\highlightcolor}{motion blur} that heavily affect SIFT-based methods.
This shows the potential of CNN-based methods. 




\subsection{Complex large-scale indoor localization}

\begin{table}[t]
  \newcommand*{\poseerr}[2]{\SI{#1}{\m}, \SI{#2}{\degree}}
  \centering
  \small
  \caption{Median localization accuracy on \textcolor{\highlightcolor}{TUM-LSI}. }\label{tab:lsi}
  	\vspace{-0.1cm}
\begin{tabular}{@{} l c c  c c @{}} \toprule
   Area  & \# train/test  & PoseNet~\cite{kendall2015posenet} & Proposed \\ \midrule
   \SI{5575}{\meter\squared} &  875/220 &  \poseerr{1.87}{6.14} & \poseerr{1.31}{2.79} (30,55)\\
  \bottomrule
\end{tabular}
	\vspace{-0.3cm}
\end{table}

\begin{figure}[t]
	\centering
	\begin{subfigure}{0.8\linewidth}
	\centering
	\includegraphics[height=3cm]{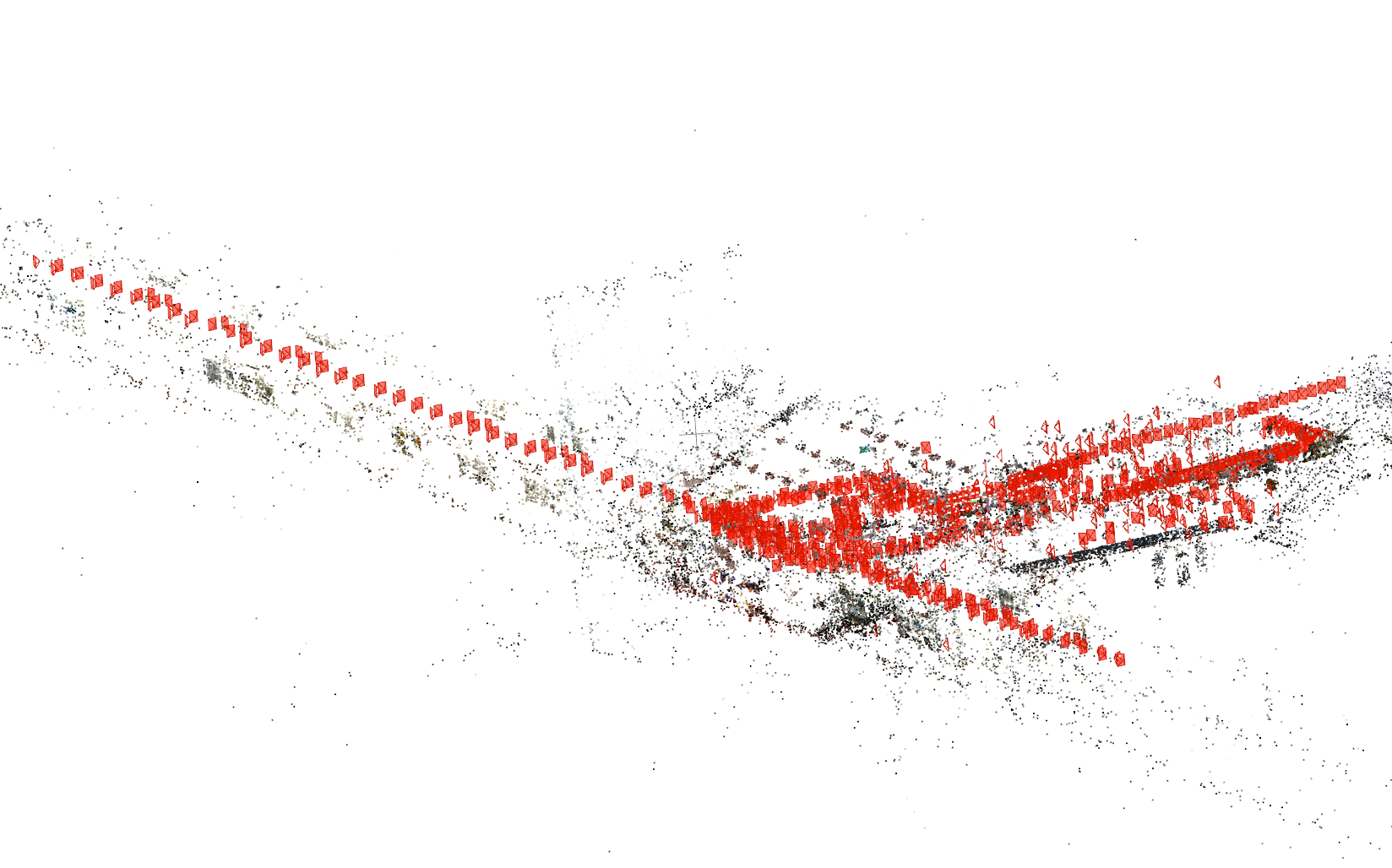}
	\end{subfigure}\\
	\begin{subfigure}{0.8\linewidth}
	\centering
	\includegraphics[height=3cm]{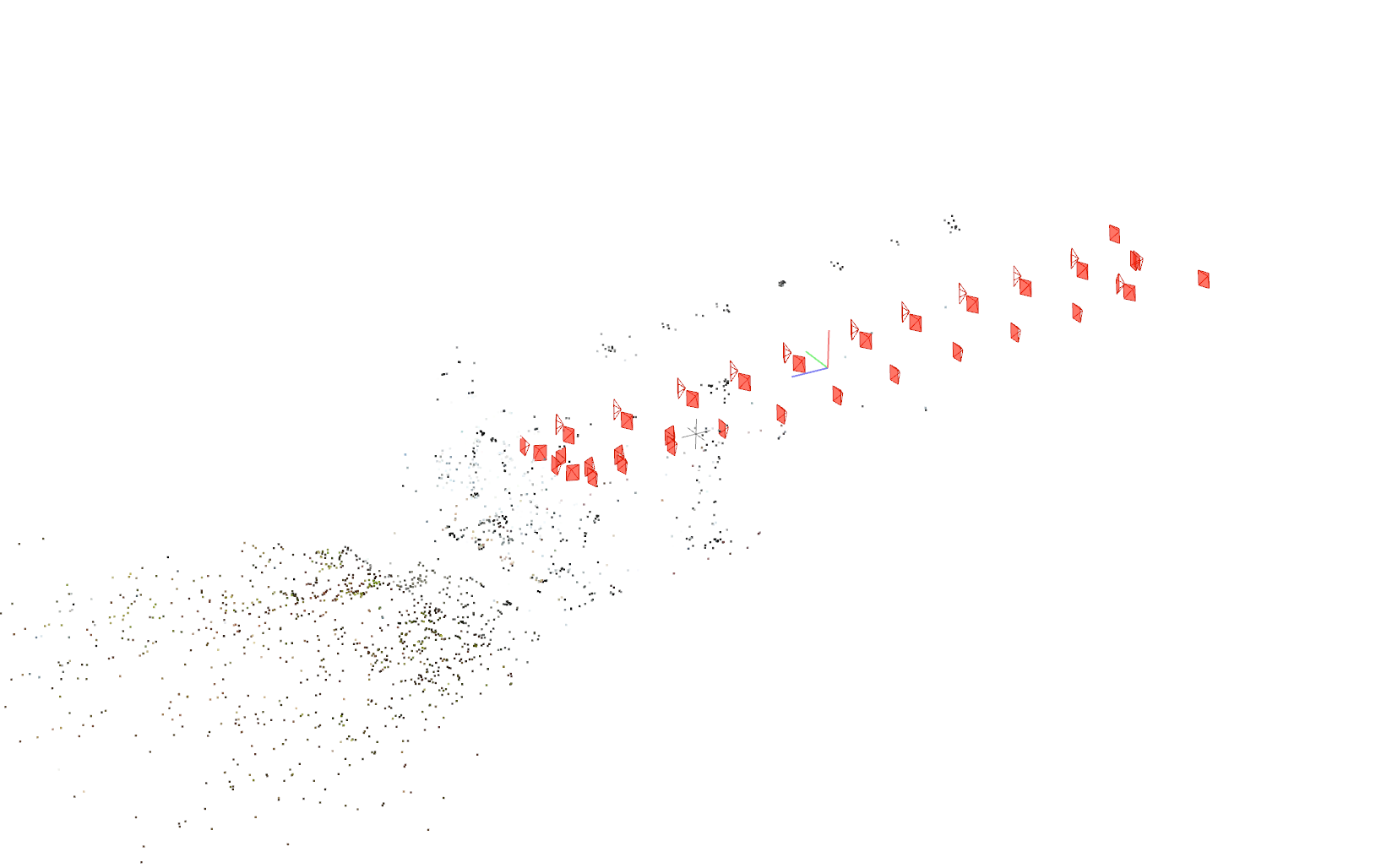}
	\end{subfigure}%
	\caption{Failed \textcolor{\highlightcolor}{SfM} reconstruction\textcolor{\highlightcolor}{s} for the \textcolor{\highlightcolor}{TUM-LSI  dataset, obtained with COLMAP}. The first reconstruction contains two stairwells collapsed into one due to repetitive structures. Due to a lack of sufficient matches, the method was unable to connect a sequence of images and therefore creates a second separate \textcolor{\highlightcolor}{model} of one of the hallways.} 
		\vspace{-0.2cm}
	\label{fig:reconstruction_fail}
\end{figure}


In our last experiment, we present results on the new \textcolor{\highlightcolor}{TUM-LSI dataset}. It covers a total area of 5,575 \textcolor{\highlightcolor}{$m^2$}, the same order of magnitude as the outdoor localization dataset, and much larger than typical indoor datasets like 7Scenes. 

Figure~\ref{fig:lsi-example-images} shows an example from the dataset that contains large textureless surfaces. 
These surfaces are known to cause problems for methods based on local features. 
In fact, we were not able to obtain correct SfM reconstructions for the \textcolor{\highlightcolor}{TUM-LSI} dataset. 
The lack of texture in most parts of the images, combined with repetitive scene elements, causes both VisualSFM and COLMAP to fold repetitive structures onto themselves. 
For example, the two separate stairwells (red floor in Figure~\ref{fig:lsi-example-images}) are mistaken for a single stairwell (\cf Figure~\ref{fig:reconstruction_fail}). 
\textcolor{\highlightcolor}{As the resulting 3D model does not reflect the true 3D structure of the scene, there is no point in applying Active Search or any other SIFT-based method. Notice that s}uch repetitive structures \textcolor{\highlightcolor}{would} cause Active Search to fail even if a good model is provided.

For the experiments on the \textcolor{\highlightcolor}{TUM-LSI dataset}, we ignore the ceiling-facing cameras. As we can see in Table~\ref{tab:lsi}, our method outperforms PoseNet~\cite{kendall2015posenet} by almost 30\% in positional error and 55\% orientation error, showing a similar improvement as for other datasets. 
To the best of our knowledge, 
 we are the first to showcase a scenario where CNN-based methods succeed while SIFT-based approaches fail. 
On this challenging sequence,  
our method achieves an average error of around 1m. 
In our opinion, this demonstrates that CNN-based methods are indeed a promising avenue to tackle hard localization problems such as 
repetitive structures and textureless walls, which are predominant in modern buildings, and are a problem for classic SIFT-based localization methods.






\section{Conclusion}

In this paper, we address the challenge of image-based localization of a camera or an autonomous system with a novel deep learning architecture that combines a CNN with LSTM units.  Rather than precomputing feature points and building 
a map as done in traditional SIFT-based localization techniques, we determine a direct mapping from input image to camera pose.  
With a systematic evaluation on existing indoor and outdoor datasets, we show that our LSTM-based structured feature correlation can lead to drastic improvements in localization performance  
compared to \textcolor{\highlightcolor}{other} CNN-based methods. 
Furthermore, we are the first to show a comparison of SIFT-based \textcolor{\highlightcolor}{and} CNN-based localization methods, showing that classic SIFT approaches still outperform all published CNN-based methods to date 
on standard benchmark datasets. To answer the ensuing question whether CNN-based localization is a promising direction of research, we demonstrate that our approach succeeds in a very challenging scenario where SIFT-based methods fail. To this end,  
we introduce a new challenging large-scale indoor sequence with accurate ground truth. 
Besides aiming to close the gap in accuracy between SIFT- and CNN-based methods, we believe that exploring CNN-based localization in hard scenarios is a promising research direction. 
{
\small
\PAR{Acknowledgements.} This work was partially funded by the ERC Consolidator grant \emph{3D Reloaded} and a Sofja Kovalevskaja Award from the Alexander von Humboldt Foundation, endowed by the Federal Ministry of Education and Research.
}

{\small
\bibliographystyle{ieee}
\bibliography{poseiccv}
}

\end{document}